\documentclass[10pt,twocolumn,letterpaper]{article}
\usepackage{iccv}
\usepackage{times}
\usepackage{epsfig}
\usepackage{graphicx}
\usepackage{amsmath}
\usepackage{amssymb}
\usepackage{multirow}
\usepackage{float}
\usepackage{soul}   
\usepackage{booktabs}

\newcommand{\nbf}[1]{{\noindent\textbf{#1.}}}

\usepackage[breaklinks=true,bookmarks=false]{hyperref}

\iccvfinalcopy 
\def\iccvPaperID{****} 

\ificcvfinal\fi

\begin{document}

\title{Target-Grounded Graph-Aware Transformer for \\ Aerial Vision-and-Dialog Navigation}

\author{
	Yifei Su$^{1,2}$ ~~ Dong An$^{2}$ ~~ Yuan Xu$^{1,2}$ ~~ Kehan Chen$^{1,2}$ ~~ Yan Huang$^{2}$ \\
	\vspace{-8pt}{\small~}\\
	$^{1}$School of Artificial Intelligence, UCAS ~~$^{2}$Institute of Automation, Chinese Academy of Sciences \\
	\vspace{-10pt}
	{\tt\small suyifei2022@ia.ac.cn}
}

\maketitle
\ificcvfinal\thispagestyle{empty}\fi

\begin{abstract}
    This report details the methods of the winning entry of the AVDN Challenge in ICCV CLVL 2023. 
    The competition addresses the Aerial Navigation from Dialog History (ANDH) task, which requires a drone agent to associate dialog history with aerial observations to reach the destination. For better cross-modal grounding abilities of the drone agent, we propose a Target-Grounded Graph-Aware Transformer (TG-GAT) framework. Concretely, TG-GAT first leverages a graph-aware transformer to capture spatiotemporal dependency, which benefits navigation state tracking and robust action planning. 
    In addition, an auxiliary visual grounding task is devised to boost the agent's awareness of referred landmarks. 
    Moreover, a hybrid augmentation strategy based on large language models is utilized to mitigate data scarcity limitations.
    Our TG-GAT framework won the AVDN Challenge, with 2.2\% and 3.0\% absolute improvements over the baseline on \texttt{SPL} and \texttt{SR} metrics, respectively. The code is available at \
    \href{https://github.com/yifeisu/TG-GAT}{https://github.com/yifeisu/TG-GAT}.
\end{abstract}


\section{Introduction}
Languag-guided mobile navigation has made significant progress in recent years. 
Numerous tasks~\cite{anderson2018vision,thomason:corl19,reverie,fan2023r2h} and methods~\cite{wang2018look,wang2019reinforced,wang2020environment,wang2020vision,hong2021vln,chen2021hamt,Chen_2022_DUET,an2022bevbert,wang2023scaling} have been proposed to solve this problem.
Recently, Fan \etal~\cite{fan2022aerial} exposed this problem in a drone scenario, which provides wide potential applications, such as food delivery and wilderness rescue.
The proposed Aerial Navigation from Dialog History (ANDH) task requires a drone agent to interpret dialog history in bird's-eye-view observations~\cite{lam2018xview} and reach the referred goal areas. 

Cross-modal grounding is a widespread challenge for language-guided navigation, while ANDH makes it even more challenging due to much longer trajectories and a wider field of view. 
This not only increases the difficulty for navigation state tracking but also precise landmark grounding from redundant observations.
Thus, Fan \etal~\cite{fan2022aerial} benchmark ANDH using a history-aware transformer~\cite{chen2021hamt,pashevich2021episodic} with human attention.
It stacks historical observations to capture long-horizon dependency and supervises visual perception with human attention masks.

Despite the progress, we found this solution still has drawbacks in three aspects. 
First, the stacked visual histories are unstructured, which may hinder the agent from comprehending directional dependency, such as ``proceed forward direction and turn right''. 
Second, the agent lacks fine-grained grounding abilities and therefore is unaware of the mentioned landmarks, leading to a sub-optimal stop policy, such as ``destination is a condominium with a sports court''.
Third, the data scarcity issue limits the agent's generalization ability to unseen environments. 

\begin{figure*}[!htbp]
	\begin{center}
		\includegraphics[width=0.98\linewidth]{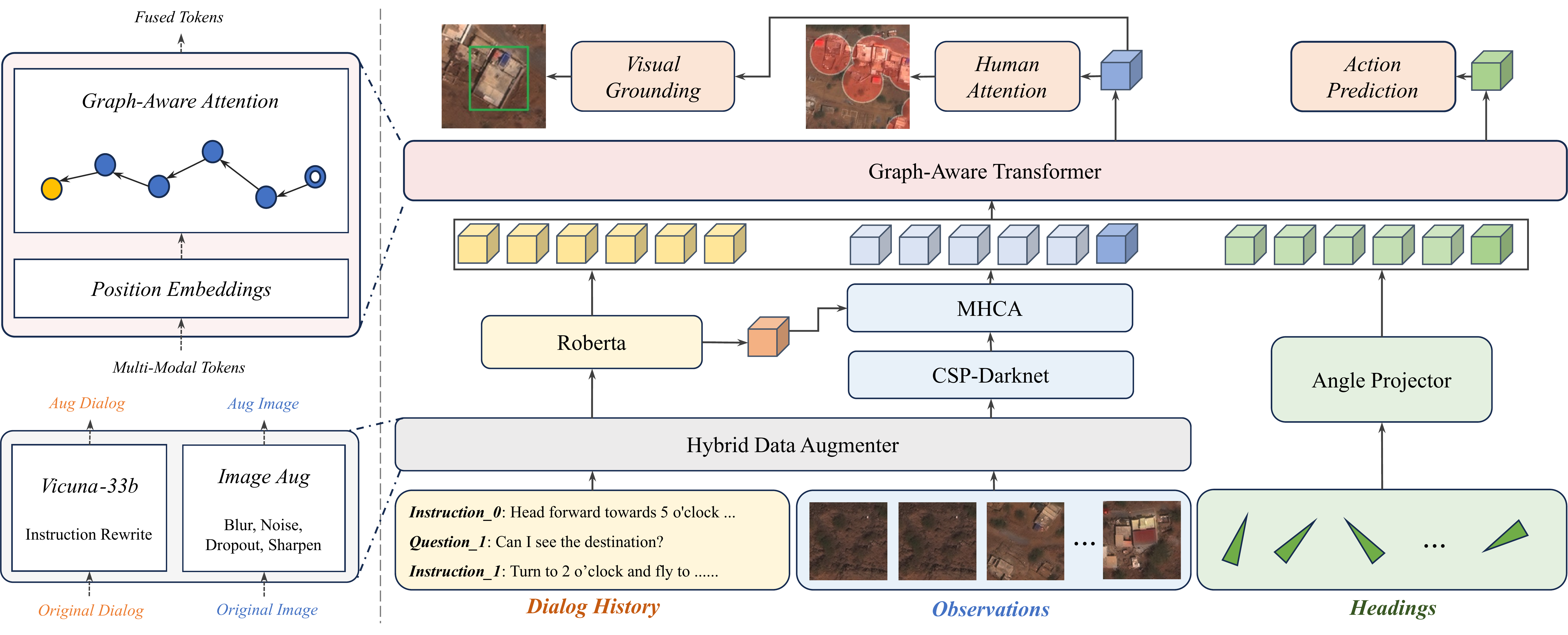}
	\end{center}
	\vspace{-3mm}
	\caption{Overview of the proposed TG-GAT framework.}
	\label{fig:overview}
\end{figure*}

To address the above challenges, we propose a Target-Grounded Graph-Aware Transformer (TG-GAT) to enable structured memory modeling and fine-grained landmark grounding. 
As illustrated in Figure~\ref{fig:overview}, TG-GAT mainly comprises three innovations: graph-aware transformer, auxiliary grounding task, and hybrid data augmenter.
The graph-aware transformer leverages a graph-attention mechanism to associate dialog with structured historical observations, providing more comprehensive spatiotemporal information for action planning. 
Beyond the original human attention supervision~\cite{fan2022aerial},  we propose a fine-grained visual grounding task for model training. 
This task can boost the agent's awareness of landmarks by forcing it to predict the precise bounding box of the referred landmark.
Moreover, to solve the data scarcity issue, the augmenter performs various data augmentation on both dialogs and observations. Specifically, a large language model~\cite{zheng2023judging} is prompted to rewrite and synthesize more human instructions, while various augmentation strategies are applied over images, \eg, image blur, random noise, pixel dropout.

Extensive experiments demonstrate the effectiveness of the proposed method. 
TG-GAT won the ICCV CLVL 2023 AVDN challenge and clearly improved the baseline model, \eg, on the test unseen split, Success weighted by inverse Path Length (\texttt{SPL}) increases from 12.9 to 15.1 and Success Rate (\texttt{SR}) increases from 15.7 to 18.7.

\section{Method}
\label{Method}
\subsection{Task Setup}
Given past dialog history, the ANDH task~\cite{fan2022aerial} requires a drone agent to accomplish a sub-trajectory during a dialog round. 
Specifically, for each episode, the agent gets the current dialog $\mathcal{D}_c=\{Q_n,I_n\}$ and historical dialogs $\mathcal{D}_h=\{I_0, Q_1,I_1, ...,Q_{n-1},I_{n-1}\}$, where $Q_*$ are questions posed by the drone, and $I_*$ are human instructions. 
The ground-truth sud-trajectory comprises a sequence of view areas denoted as $\mathcal{T} = \{\mathcal{C}_1^{gt}, \mathcal{C}_2^{gt}, ...\mathcal{C}_L^{gt}\}$, where $L$ is path length and $\mathcal{C}_*^{gt}$ represent the GPS coordinates of the $4$ view vertices. 
The final view area $\mathcal{C}_L^{gt}$ is regarded as the target region.
At each step $t$, the agent perceives the current bird's-eye-view RGB observation $V_t$ and its view corners $\mathcal{C}_t$. 
The agent also receives the current position $\mathcal{P}_t=\{x_t, y_t, z_t\}$ and heading angle $\theta_t$. 
The agent's object is to align the dialog with visual observations and reach the target $\mathcal{C}_L^{gt}$ through action $a_t=\{\Delta x_t, \Delta y_t,\Delta z_t\}$.
An episode is successful if the IoU between the final predicted view area $\mathcal{C}_{f}$ and the target $\mathcal{C}_L^{gt}$ is greater than $0.4$.

\subsection{Method Overview}
As illustrated in Figure~\ref{fig:overview}, our TG-GAT  mainly comprises three innovations: graph-aware transformer (\S\ref{mmencoding}), auxiliary grounding task (\S\ref{visgr}), and hybrid data augmenter (\S\ref{dataaug}). 
At each step $t$, the augmenter synthesizes new observation images and dialog instructions. After that, the embeddings of the three modal inputs (\ie, the dailogs, history views, and history heading angles) are acquired via unimodal encoders, and then stored in memory buffers. 
Subsequently, three types of embeddings are simultaneously fed into the graph-aware transformer to predict the next action. 
Meanwhile, the visual grounding and human attention prediction auxiliary tasks are applied for model training. We detail the training phase in \S\ref{training}.

\subsection{Multimodal Encoding}
\label{mmencoding}
We employ a pretrained Roberta~\cite{liu2019roberta} for text encoding. 
Specifically, dialog inputs $[\mathcal{D}_c, \mathcal{D}_h]$ are first tokenized and padded with special tokens [QUE], [INS] before each question and instruction following~\cite{fan2022aerial}. 
Then, they are added with position embeddings~\cite{kenton2019bert} and fed into the Roberta model to obtain contextual text embeddings.

For image encoding, the current observation $V_t$ is fed into an xView-pretrained Yolov5-x\footnote{\href{https://huggingface.co/deprem-ml/Binafarktespit-yolo5x-v1-xview}{https://huggingface.co/deprem-ml/Binafarktespit-yolo5x-v1-xview}} to extract grid features $F_t$. 
After that, we flatten $F_t$ and leverage a multi-head cross attention mechanism (MHCA)~\cite{vaswani2017attention} to aggregate it as a feature vector $\tilde{F}_t$:
{\small
\begin{equation}
\begin{aligned}
    \tilde{F}_t&=\mathrm{FFN}\left(I_\mathrm{cls}+\mathrm{MHCA}(F_t)\right), \\
    \mathrm{MHCA}(F_t)&=\mathrm{Softmax}\left(\frac{I_\mathrm{cls}W_q(F_tW_k) ^T}{\sqrt{d}}\right)F_tW_v
\end{aligned}
\vspace{-1mm}
\end{equation}
}%
where $I_\mathrm{cls}$ is the contextual embedding of the [CLS] token, and $W_q, W_k, W_v$ are three learnable matrices.
Meanwhile, the sine and cosine encoding of heading angles are fed into a three-layer dense network to generate direction embeddings.\textbf{}
Then, we store the image and direction embeddings in memory buffers similar to~\cite{fan2022aerial}.

\begin{figure}[tb]
	\begin{center}
		\includegraphics[width=0.98\linewidth]{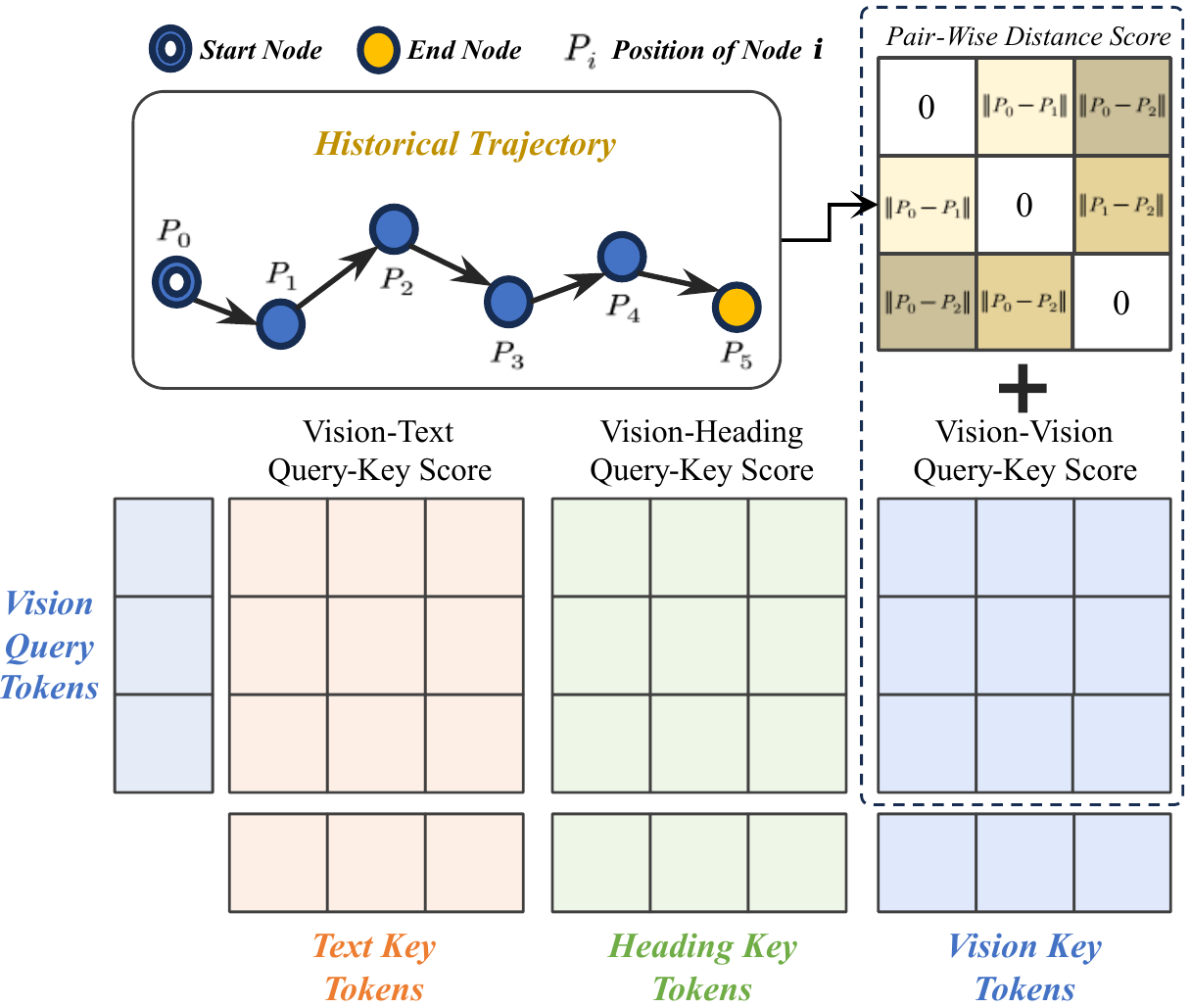}
	\end{center}
        \vspace{-3mm}
	\caption{GAT combines standard query-key similarity (bottom) and pair-wise distance similarity as final attention scores.}
        \vspace{-3mm}
	\label{gaet}
\end{figure}

Subsequently, we acquire all historical embeddings from the memory buffers and perform multimodal encoding. 
The obtained image and direction embeddings are added with step encodings~\cite{fan2022aerial}, concatenated with text embeddings and then fed into a Graph-Aware Transformer (GAT). 
Inspired by DUET~\cite{Chen_2022_DUET}, GAT injects structure information into multimodal memory encoding for better spatial dependency modeling. 
Concretely, GAT replaces the traditional self-attention~\cite{vaswani2017attention} with a graph-attention mechanism. As shown in Figure \ref{gaet}, besides the standard query-key attention similarity, GAT introduces an extra distance similarity $G$ computed as follows:
\begin{equation}
    \label{gga}
	G = W_e E + b_e
\end{equation}
where $E$ is the pair-wise distance matrix of historical locations, and $W_e, b_e$ are two learnable parameters. 
In particular, the distance $E_{ij}$ between any two historical locations $\mathcal{P}_i$ and $\mathcal{P}_j$ is computed through $\left\| [x_i,y_i] - [x_j, y_j] \right\|_2 $. 

As for the auxiliary tasks and action prediction, akin to \cite{fan2022aerial}, we employ the multimodel embedding of the current heading angle to predict the next action. 
Meanwhile, the multimodel encoding of the current observation is utilized for human attention prediction~\cite{fan2022aerial} and cross-model grounding (\S\ref{visgr}).

\subsection{Visual Grounding Task}
\label{visgr} 
To enhance the agent's landmark grounding capability, we devise an auxiliary visual grounding task to predict the bounding box of the referred destination.
Specifically, the multimodal embedding of current observation is fed into a three-layer dense network to predict the $1$-dim confidence $\hat{c}$ and $4$-dim box coordinates of destination $\hat{b}=[ \hat{x},\hat{y},\hat{w},\hat{h} ]$, where $\hat{c}$ represents the likelihood that the current observation contains the target area. 

Same as TransVG~\cite{deng2021transvg}, we leverage the smooth-L1 loss $\mathcal{L}_\mathrm{l1}$ and GIoU loss $\mathcal{L}_\mathrm{giou}$ to supervise the box regression, along with binary cross-entropy loss $\mathcal{L}_\mathrm{bce}$ to optimize confidence prediction. The overall cross-modal grounding loss $\mathcal{L}_\mathrm{gr}$ is obtained as follows:
{\small
    \begin{equation}
        \mathcal{L}_\mathrm{gr} = \kappa_1\mathcal{L}_\mathrm{l1}(\hat{b}, b)+\kappa_2\mathcal{L}_\mathrm{giou}(\hat{b}, b)+\kappa_3\mathcal{L}_\mathrm{bce}(\hat{c}, c)
    \end{equation}
}%
Where $b=[{x},{y},{w},{h}]$ and $c$ denote the ground-truth bounding box and confidence, and $c$ is set to $0$ when the target area is outside the current observation. $\kappa_*$ are the weight coefficients to balance the three terms.

\vspace{1 pt}
\subsection{Augmentation Strategy}
\label{dataaug}
The ANDH dataset only contains 4951 dialog-trajectory pairs for training, which can be insufficient to learn a robust policy. 
To this end, we propose a hybrid data augmenter for extra regularization. 
On the one hand, the augmenter implements various image augmentations during training via Albumentations~\cite{info11020125}, \eg, image blur, random noise, random contrast, pixel dropout. 
On the other hand, we prompt a large language model Vicuna-33b~\cite{zheng2023judging} to synthesize more human instructions. 
The used prompt template is ``\textit{Rewrite the given sentence in 5 different ways and keep the details unchanged.}"
Given an original instruction, ``\textit{Hi drone, head southwest and pass over a building, and your destination is the small green building.}", the generated sentence is like ``\textit{Drone, navigate southwest and cross over a building; your goal is the little green building.}". In this way, we scale the ANDH instructions to 5$\times$ larger than the origin.

\subsection{Training}
\label{training}
Similar to~\cite{fan2022aerial}, we alternatively run teacher-forcing and student-forcing learning to train our model. 
The model is optimized via a multi-task loss as follows:
{\small
\begin{equation}
	\mathcal{L}_\mathrm{sum} = \lambda_1\mathcal{L}_\mathrm{nav} + \lambda_2\mathcal{L}_\mathrm{hap} + \lambda_3\mathcal{L}_\mathrm{gr}
\end{equation}
}%
where the $\mathcal{L}_{\mathrm{nav}}$ and $\mathcal{L}_{\mathrm{hap}}$ are the navigation prediction loss and human attention loss used in~\cite{fan2022aerial}. $\mathcal{L}_{\mathrm{gr}}$ denotes the proposed visual grounding auxiliary task. $\lambda\mathrm{*}$ are the loss weight coefficients to balance the three terms.

\section{Experiments}
\label{Experiments}
\subsection{Dataset and Evaluation Metrics}
\label{data and metric}
The ANDH dataset used in this competition contains 6,269 dialog-trajectory pairs which are divided into training, seen validation, unseen validation, and unseen testing splits.
More attention is paid to the performance in unseen environments, and the unseen testing split is used for leaderboard evaluation. 
We use standard metrics~\cite{fan2022aerial} to measure navigation performance, i.e., \textit{Success Rate} (\texttt{SR}): number of the predicted trajectory being regarded as successful; \textit{Success weighted by inverse Path Length} (\texttt{SPL}): \texttt{SR} weighted by the total length of the navigation trajectory; \textit{Goal Progress} (\texttt{GP}): distance of the navigation progress made towards the destination area. \texttt{SPL} is the primary metric that considers both
navigation accuracy and efficiency, while \texttt{SR} and \texttt{GP} are the main secondary metrics.

\subsection{Implementation Details}
\label{implementation details}
The number of layers for the MHCA, text encoder, and graph-aware transformer is set as 1, 9, and 2, respectively. 
The hidden size of text embeddings, image embeddings, and angle embeddings are uniformly set to 768.
For cross-modal grounding loss, we empirically set $\kappa_1, \kappa_2, \kappa_3$ to $1, 3, 1.5$ in our experiments.Additionally, the $\lambda_1, \lambda_2, \lambda_3$ are set to $0.2, 0.1, 0.25$ for multi-task learning. 
During training, image augmentation is implemented with a certain probability $p=0.4$, while the original observations are used for inference.
Our experiments are conducted on 2 NVIDIA RTX 3090 GPUs. All models are optimized for 200,000 iterations with a batch size of 4 and a learning rate of 1e-5 using AdamW optimizer. All parameters, including Roberta and Yolov5-x, are fine-tuned during the training process same as~\cite{fan2022aerial}. 
The optimal checkpoint is determined by the best performance on unseen validation split.

\begin{table}[t]
\centering
\resizebox{0.41\textwidth}{!}{
\begin{tabular}{cccccc}
\toprule
\multirow{2}{*}{Model} & \multirow{2}{*}{\begin{tabular}[c]{@{}c@{}}Image\\ Aug\end{tabular}} & \multirow{2}{*}{\begin{tabular}[c]{@{}c@{}}Instruction \\ Aug\end{tabular}} & \multicolumn{3}{c}{Unseen Validation} \\
\cmidrule{4-6} &  &  & SPL$\uparrow$ & SR$\uparrow$ & GP$\uparrow$ \\ 
\midrule
baseline & $\times$   & $\times$          &  16.5   &  20.4  &  55.2  \\
1 & \checkmark & $\times$     &  16.9   &  20.2  &  51.9 \\
2 & \checkmark &  \checkmark   &  \textbf{18.2} & \textbf{20.9} & \textbf{58.2} \\
\bottomrule
\end{tabular}%
}
\vspace{2pt}
\caption{The effect of data augmentation.}
\label{com_aug}
\end{table}

\begin{table}[t]
\centering
\resizebox{0.45\textwidth}{!}{%
\begin{tabular}{ccccccc}
\toprule
\multirow{2}{*}{Model} &
\multirow{2}{*}{\begin{tabular}[c]{@{}c@{}}Image\\ Aug\end{tabular}} &
\multirow{2}{*}{MHCA} &
\multirow{2}{*}{\begin{tabular}[c]{@{}c@{}}Grounding\\ Task\end{tabular}} &
\multicolumn{3}{c}{Unseen Validation} \\
\cmidrule{5-7} &  &  &  & SPL$\uparrow$ & SR$\uparrow$ & GP$\uparrow$ \\ 
\midrule
1 & \checkmark& $\times$ & $\times$ & 16.9 & 20.2 & 51.9 \\
2 & \checkmark & \checkmark & $\times$ & 18.1 & \textbf{22.4} & \textbf{58.9} \\
3 & \checkmark & \checkmark & \checkmark & \textbf{18.9} & 22.2 & 56.0 \\ 
\bottomrule
\end{tabular}%
}
\vspace{2pt}
\caption{The effect of MHCA and grounding task.}
\vspace{-3mm}
\label{com_gr}
\end{table}

\subsection{Ablation Study}
This section presents ablation experiments to evaluate specific components of the proposed TG-GAT. All results are reported on the unseen validation split.

\nbf{Analysis of data augmentation strategy}
Table~\ref{com_aug} presents the effect of the hybrid data augmenter. 
The baseline is the HAA-Transformer model reported in~\cite{fan2022aerial}. 
Model 1 enhances the baseline by integrating the image augmentation, resulting in a 0.4 \texttt{SPL} improvement. 
Model 2 further incorporates the instruction augmentation, leading to $\uparrow$ 1.7 \texttt{SPL}, $\uparrow$ 0.5 \texttt{SR} and $\uparrow$ 3.0 \texttt{GP} gains. 
This indicates image augmentation and instruction augmentation are complementary, and instruction diversity is crucial to train a robust ANDH agent.

\nbf{Analysis of MHCA and grounding task}
We further conduct ablation experiments on the MHCA mechanism and visual grounding task based on model 1 in Table \ref{com_aug}. In Table~\ref{com_gr}, compared with model 1, model 2 with MHCA boosts \texttt{SPL} from 16.9 to 18.1, \texttt{SR} from 20.2 to 22.4 and \texttt{GP} from 51.9 to 58.9 on unseen validation split. 
It indicates that MHCA can provide a more informative visual embedding for navigation. 
Meanwhile, model 3 achieves the best \texttt{SPL} of 18.9 with the extra visual grounding task, but \texttt{GP} suffers a 2.9 drop.
A potential reason is that the grounding task forces the drone to overlook the referred destination rather than a more appropriate location.

\nbf{Analysis of different encoders and GAT} 
Expanding upon model 3 in Table \ref{com_gr}, we further investigate the impact of multimodal encoders and GAT on model performance.
Comparing model 2 to model 1, the Roberta and xView-pretrained Yolov5-x backbone can promote \texttt{SR} from 22.2 to 23.3.
In addition, with the GAT, model 3 achieves a higher \texttt{GP} score. 
We attribute this to the captured spatial information, which helps avoid repeated visits to the same location, thus facilitating the progress toward the destination. Although incurring a 0.4 decrement in \texttt{SPL}, model 3 achieves a more comprehensive performance across all evaluation metrics on unseen validation split.

\begin{table}[t]
\centering
\resizebox{0.45\textwidth}{!}{%
\begin{tabular}{ccccccc}
\toprule
\multirow{2}{*}{Model} & 
\multirow{2}{*}{\begin{tabular}[c]{@{}c@{}}Bert\\Yolov3\end{tabular}} & 
\multirow{2}{*}{\begin{tabular}[c]{@{}c@{}}Robert\\Yolov5-x\end{tabular}} & 
\multirow{2}{*}{GAT} & 
\multicolumn{3}{c}{Unseen Validation} \\ 
\cmidrule{5-7} &  &  &  & SPL$\uparrow$ & SR$\uparrow$ & GP$\uparrow$ \\ 
\midrule
1 & \checkmark  & $\times$  & $\times$ 		& 18.9 & 22.2 & 56.0 \\
2 & $\times$    & \checkmark & $\times$ 		& 18.8 & \textbf{23.4} & 54.3 \\
3 & $\times$    & \checkmark & \checkmark	& 18.4 & 22.6 & \textbf{58.1} \\ 
\bottomrule
\end{tabular}
}
\vspace{2pt}
\caption{Ablation studies of modal encoders and GAT}
\label{com_gaet}
\end{table}

\begin{table}[t]
\resizebox{\columnwidth}{!}{%
\begin{tabular}{lcccccc}
\toprule
& \multicolumn{3}{c}{Unseen Validation} & \multicolumn{3}{c}{Unseen Testing} \\
\cmidrule(r){2-4} \cmidrule(r){5-7}
Model           & SPL$\uparrow$ & SR$\uparrow$  & GP$\uparrow$  & SPL$\uparrow$ & SR$\uparrow$ & GP$\uparrow$  \\ 
\midrule
HAA-LSTM        & 18.3        & 20.0        & 54.4        & 12.6        & 14.1       & 50.8        \\
HAA-Transformer & 16.5        & 20.4        & 55.2        & 12.9        & 15.7       & 54.2        \\
Ours          & 18.8        & 23.3        & 54.3        & \textbf{15.1} & \textbf{18.7} & \textbf{56.5} \\ 
\bottomrule
\end{tabular}%
}
\vspace{-4pt}
\caption{AVDN challenge leaderboard results.}
\vspace{-3mm}
\label{leadboard}
\end{table}

\subsection{Leaderboard Results}
\label{leaderboard results}
Table~\ref{leadboard} shows our final results in AVDN challenge 2023. Our method outperforms previous state-of-the-art through all evaluation metrics, \eg, \texttt{SR} increases from 15.7 to 18.7 and \texttt{SPL} increases from 12.9 to 15.1. 
Note that our final submission is inferred by model 2 in Table \ref{com_gaet} due to the submission time restriction.

\section{Conclusion}
\label{Conclusion}
The ICCV CLVL 2023 AVDN challenge introduces a difficult and realistic testbed to evaluate language-guided navigation in the drone scenario. 
By combining graph-aware transformer, auxiliary cross-modal grounding task and data augmentation, the proposed TG-GAT sets the new state-of-the-art, \eg, the \texttt{SPL} increases from 12.9 to 15.1, \texttt{SR} from 15.7 to 18.7, and \texttt{GP} from 54.2 to 56.5. 
Nevertheless, this performance is still far from perfect. We hope the proposed method can serve as a strong baseline for further research on this challenging task.


{
\small
\bibliographystyle{unsrt}
\bibliography{egbib}
}

\end{document}